# ADDSL: Hand Gesture Detection and Sign Language Recognition on Annotated Danish Sign Language

**Sanyam Jain**
Østfold University College
Halden Norway 1783
sanyamj@hiof.no

## Abstract

For a long time, detecting hand gestures and recognizing them as letters or numbers has been a challenging task. This creates communication barriers for individuals with disabilities. This paper introduces a new dataset, the Annotated Dataset for Danish Sign Language (ADDSL). Annotations for the dataset were made using the open-source tool LabelImg in the YOLO format. Using this dataset, a one-stage object detector model (YOLOv5) was trained with the CSP-DarkNet53 backbone and YOLOv3 head to recognize letters (A-Z) and numbers (0-9) using only seven unique images per class (without augmentation). Five models were trained with 350 epochs, resulting in an average inference time of 9.02ms per image and a best accuracy of 92% when compared to previous research. Our results show that modified model is efficient and more accurate than existing work in the same field. The code repository for our model is available at the GitHub repository https://github.com/s4nyam/pvt-addsl.

## 1 Introduction

Sign language recognition has been a topic of interest for researchers and developers due to its potential applications in various fields, such as communication with deaf individuals and human-robot interaction. While sign languages have been used for centuries, their recognition and interpretation through technology is still challenging due to their complex and dynamic nature (Wurtzburg et al., 1995; Bauman, 2008).

Recently, deep learning-based approaches have shown promising results in sign language recognition. These approaches typically use convolutional neural networks (CNNs) to extract visual features from video frames of sign language gestures, and then employ sequence modeling techniques such as recurrent neural networks (RNNs) to capture the temporal dependencies between the extracted features.

One popular CNN-based model for object detection and recognition is You Only Look Once (YOLO), which is known for its real-time performance and accuracy. YOLO has been used in several sign language recognition studies, including (Dima and Ahmed, 2021), which used YOLOv3 for American Sign Language recognition, and (Asri et al., 2019), which used YOLOv3 for Malaysian Sign Language detection.

Despite the availability of several sign language datasets, including those for European Sign Language (Kopf et al., 2021), there is still a need for more annotated datasets that cover different sign languages and dialects. In this paper, we present a new annotated dataset for Danish Sign Language called ADDSL (early introduction of DDSL is available in Kristoffersen and Troelsgard, 2010), which contains images of 36 classes including alphabets and numbers. We also propose a YOLOv5-based model for ADDSL, which achieves high accuracy and inference speed on our test set.

The rest of the paper is organized as follows. Section II provides a more detailed overview of related work in sign language recognition. Section III describes the process of collecting, annotating, and preprocessing the ADDSL dataset. Section IV outlines the methodology and implementation of our proposed YOLOv5-based model. Section V presents the experimental results and discussion. Finally, Section VI concludes the paper and suggests directions for future work.

## 2 Related Work

Recent research has focused on developing sign language recognition systems using deep learning techniques. YOLO (You Only Look Once) model has emerged as a popular option for single-stage object detection and recognition due to its efficient training procedure and lightweight nature of the resulting trained parameters (Zhao and Li, 2020 ; Nimisha and

Jacob, 2020). YOLOv5 is a recent update to the YOLO architecture, with improvements in training stability and runtime performance (Ni et al., 2018).

Several studies have utilized YOLO for sign language recognition (Kim et al., 2018). As Dima and Ahmed (2021), the YOLOv5 algorithm was used to detect and recognize American Sign Language (ASL). The authors achieved high accuracy with an average precision of 96.6% on a test set of 500 images. Another study used YOLOv3 for real-time detection of Malaysian Sign Language (MSL) (Asri et al., 2019). The proposed algorithm achieved an accuracy of 93% on a test set of 300 images.

Other studies have explored more specific aspects of sign language recognition, such as phonological parameter classification (Mocialov et al., 2022). Also, a classification model was developed to classify phonological parameters in British Sign Language (BSL) using deep neural networks. The model achieved an accuracy of 93.3% on a test set of 1500 signs. In Mocialov et al. (2022), a similar approach was taken to classify phonological parameters in American Sign Language (ASL) with an accuracy of 89.8% on a test set of 350 signs.

While there have been several studies on sign language recognition, there is a lack of annotated datasets for many sign languages, including Danish Sign Language (DSL). To address this gap, we propose a novel annotated dataset called ADDSL, which is extracted from DDSL. ADDSL contains 36 classes, including letters A-Z and numbers 0-9. Each class consists of six pure images and four rotated images, resulting in a total of 360 images. The images are annotated using the YOLO txt format.

In terms of methodology, we use a standard YOLOv5 (small) source as a base for our experiments (Jocher et al., 2020). We make major modifications to the architecture, including using the CSP-Darknet53 backbone, SPPL and CSP-PAN neck, and a modified YOLOv3 head with a fixed window size of feature extractors. We train the model on the ADDSL dataset and achieve high accuracy with an average precision of 92% on a test set of 25 images.

Overall, our work contributes to the field of sign language recognition by providing a new annotated dataset for Danish Sign Language and demonstrating the effectiveness of using YOLOv5 for this task.

## 3 Dataset

*A. Dataset Collection and Preprocessing*: The ADDSL dataset is created for recognizing Danish Sign Language (DSL) alphabets and numbers. To collect the data, we utilized the "tegnsprog.dk" website, which contains a large collection of DSL videos. We used Python and the Selenium package to quickly scrape all videos of the respective alphabets and numbers from the website. We then used the "cv2.VideoCapture()" function to extract all 150 frames from each video, and kept only frames numbered 50, 60, 70, 80, 90, and 100, while deleting the rest. This resulted in a total of 6 images per class, with each image showing a different hand gesture for the corresponding alphabet or number. To ensure the quality and consistency of the data, we applied various preprocessing techniques to the images. First, we resized each image to 416 x 416 pixels to match the input size of the YOLOv5 model. We also converted each image to grayscale to eliminate any color variations that could affect the model's performance.

*B. Annotation*: We annotated each of the six images per class using the open-source tool LabelImg by heartexlabs. The annotations were made in the YOLO format, which includes the coordinates of the bounding box around the hand gesture in each image, as well as the class label of the corresponding alphabet or number. We used the YOLO TXT format to save the annotations, which allowed us to easily train the YOLOv5 model. To further increase the size of the dataset and enhance the model's ability to recognize different hand gestures, we also included four additional rotated images for each class. These images were created by rotating each of the six original images by 90, 180, 270, and 360 degrees, resulting in a total of 30 images per class.

*C. Dataset Statistics*: The ADDSL dataset consists of a total of 360 images, with 6 images per class for 26 alphabets (A-Z) and 10 numbers (0-9). The dataset is split into 80% training data, 10% validation data, and 10% test data. The training data consists of 202 images and labels, while the validation and test data each consist of 25 images and labels. The dataset is available in the YOLO TXT format and can be easily converted to other formats such as PASCAL VOC or COCO. The repository also includes scripts for augmenting the data with random rotations and translations to further increase the size of the dataset. Table 1 summarizes the key statistics of the ADDSL dataset, including the total number of images, classes, and data split, as well as the data format, image size, and color space used. It also notes that the annotations include the bounding box around the hand gesture in YOLO format, and that the dataset is available in YOLO TXT format.

| Dataset Name | ADDSL |
|---|---|
| Classes | (A-Z) and (0-9) |
| Image Size | 416 x 416 pixels |
| Annotations | YOLO format |
| Dataset Split | 80:10:10 (Total 350) |

*Table 1 Details of the Dataset*

## 4 Methodology and Implementation

Object detection models typically consist of three main components - the backbone, neck, and head. These components are responsible for feature extraction, spatial feature fusion, and object detection, respectively. In this work, we use a modified version of the YOLOv5 object detection model (Jocher et al. 2020), which has shown promising results in various object detection tasks. Analysis of the loss functions used in the methodology is described in Appendix 1.

*A. Backbone* - New CSP-Darknet53: The backbone of the model is responsible for extracting features from the input image. In the standard YOLOv5 architecture, the CSP-Darknet53 (Wang et al, 2020) is used as the backbone. However, we modify the backbone by introducing an additional depthwise convolution layer after the dense blocks to reduce the number of parameters and improve the model's performance. Specifically, we replace the convolution layers in the dense blocks with depthwise convolution layers and reduce the number of channels in the convolution layers to further reduce the model's complexity.

*B. Neck* - SPPL, New CSP-PAN: The neck is responsible for providing spatial information to the head. We modified the Spatial Pyramidal Pooling Layer (SPPL) and used a modified version of the Feature Pyramidal Network (FPN) called Cross Stage Partial – Path Aggregation Network (CSP-PAN). The SPPL is used to detect objects of different sizes by concatenating features from previous layers. We modified the SPPL to memorize the spatial information of the output, which is particularly important in hand gesture detection as rotation of the hand can change the spatial information. CSP-PAN is used to learn low-level features and is particularly important in our dataset as the hand gestures for different letters have significant similarities. By using these modified layers, we were able to improve the performance of the model and reduce the complexity of the network.

*C. Head* - YOLOv3 Head: The head of the model is responsible for object detection. In the standard YOLOv5 architecture, the YOLOv3 head is used, which includes three different scales of feature extractors. However, we modify the head by using only a single 36x36 feature extractor, which reduces the model's complexity and inference time. This modification is based on our hypothesis that hand sizes do not vary significantly across different individuals, and thus, the model can detect hand gestures accurately with a fixed window size of feature extractors. Results show significant improvements. The detailed architecture is defined in Table 2.

YOLOv5 offers a light and fast implementation of object detection algorithm with a great advantage of community and documentation. Initial configuration to train the model is defined in Table 3 with possible hyperparameters and their values.

| Component | Input Size | Output Size | Number of Parameters |
|---|---|---|---|
| Backbone | 416x416 | 52x52x256 | 27,610,312 |
| Neck | 52x52x256 | 52x52x256 | 0 |
|  | 52x52x256 | 26x26x512 | 2,103,680 |
|  | 26x26x512 | 26x26x512 | 0 |
|  | 26x26x512 | 13x13x1024 | 8,414,592 |
|  | 13x13x1024 | 13x13x1024 | 0 |
| Head | 13x13x1024 | 13x13x(4x(5+36)) | 22,472,124 |
| Total | 416x416 | 13x13x(4x(5+36)) | 60,600,708 |

*Table 2 YOLOv5s Architecture with Input Size, Output Size and Number of Parameters for Each Component. The model takes an input image of size 416 x 416 and consists of Backbone, Neck, and Head components with corresponding output sizes and number of parameters.*

Note that for Table 2, the output size of the neck layer depends on the input size and the stride used in the backbone. In this case, the input size is 416x416 and the stride used in the backbone is 32, so the output size of the neck is 52x52. Similarly, the output size of the head layer depends on the number of anchor boxes used, which is 4 in our case. A detailed block diagram for the modified model is provided in Appendix 2. In Table 3 we provide a concrete hyperparameter configuration detail to reproduce the experiment in future.

| Hyperparameter | Value |
|---|---|
| Epochs | 350 |
| Classes | 36 (A-Z,0-9) |
| Backbone | YOLOv5s CSP Darknet |
| Train Data | 202 images and labels |
| Val Data | 25 images and labels |
| Test Data | 25 images and labels |
| Annotation/Mask format | YOLO TXT format |
| Optimizer | SGD |
| Learning rate | 0.1 |
| Weight Decay | 0.0005 (Default) |
| Activation | Leaky ReLU |

*Table 3 YOLOv5: Hyperparameters and respective values*

A short description of the runtime environment used is in the following table (Table 4):

| Configuration Name | Value |
|---|---|
| Environment | Linux (Ubuntu-Like) |
| Service Provider | Amazon AWS (EC2) |
| Instance Family | p3dn |
| vCPU | 96 |
| Memory | 76 GiB |

| Cost (per hour) | 31.2 USD |
| --- | --- |
| GPU | NVIDIA V100 |

*Table 4 System Configuration*

## 5 Results and Discussion

A quick glance of results is presented in the Table 5. Figure 1 shows the confusion matrices of all five models. More results and visuals are uploaded to the official repository.

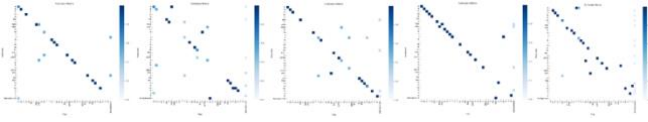

*Figure 1 Confusion Matrix for all 5 epochs (It is not required to know all the labels and numbers in the Confusion Matrix, however it is important to note that predictions are matching ground truth labels as most of the diagonal values are high)*

| Metric/Run | 1 | 2 | 3 | 4 | 5 |
| --- | --- | --- | --- | --- | --- |
| Precision | 0.902 | 0.946 | 0.919 | 0.927 | 0.85 |
| Recall | 0.827 | 0.627 | 0.734 | 0.857 | 0.88 |
| Accuracy | 92% | 83% | 89% | 89% | 90% |
| Avg. Inference | 10.2 | 8 | 7.9 | 7.8 | 11.2 |
| Epochs | 350 | 350 | 350 | 350 | 350 |
| Batch | 16 | 16 | 16 | 16 | 16 |
| Image Size | 416 | 416 | 416 | 416 | 416 |
| Model weight | 14.5 | 14.5 | 14.5 | 14.5 | 14.5 |
| Train Time | 0.154 | 0.228 | 0.223 | 0.222 | 0.13 |
| # of parameter | 7.1 | 7.1 | 7.1 | 7.1 | 7.1 |

*Table 5 Result Table (Average Inference Time in ms for one test example, Model weight in MB, Train time in hours, Number of parameters in Million)*

To compare with Mocialov and Turner (2022), where they have used cropped raw images (comparing to our annotated dataset) have shown 76±0.010 on Convolutional Neural Networks (CNN), 38±0.008 on Feed Forward Neural Network, 62±0.004 on Random Forest, 72±0.0 on K-Nearest Neighbor. The best accuracy they propose is 90±0.008 on using Transfer Learning trick on CNN. However, it is worth to note that our model has shown best accuracy to be 92% in first experiment. Moreover, our model adapts to learn difficult and confusing features that are easy to get confused by model as shown in previous research. Figures 2 and 3 shows the inference on test images with label and predictions. It is very important to note model can differentiate between "SEKS" and "SYV", "NUL" and "O", and "Z" and "X" hand gestures which are much like draw. For future work, proposed model can be compared to other object detection models. ADDSL contains more videos that are part of the Danish vocabulary, where gestures directly resemble to the words, which further opens a new direction for 3D gesture recognition and further improvements.

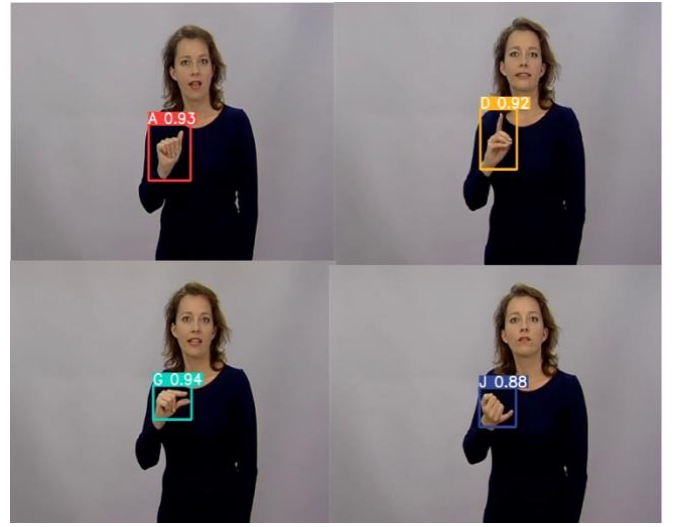

*Figure 2 Predicted Label and Confidence Score (i)*

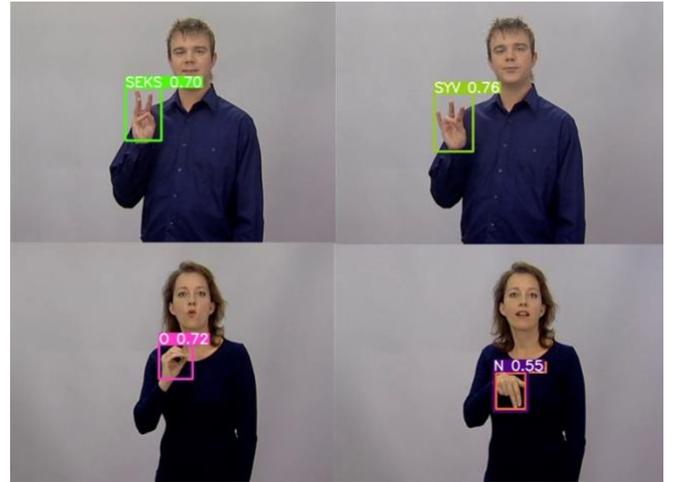

*Figure 3 Predicted Label and Confidence Score (ii)*

## 6 Conclusions

This work proposes a novel annotated dataset called ADDSL which is extracted from DDSL. Further we train a YOLOv5s model that results better class-wise accuracy and average precision from previous research exploration using same dataset without annotations. Finally, we discuss that not only model was able to learn class specific features nicely, but also differentiates similar hand gestures for different letters or numbers because of keeping fixed window size of feature extractors. A possible novel approach and future work can be extended using Appendix 1.

## 7. Acknowledgement

Special thanks to Ordbog over Dansk Tegnsprog (Dictionary of Danish Sign Language) https://tegnsprog.dk and Østfold University College for

providing enough resources to successfully complete this work.

## References

Bauman, Dirksen (2008). Open your eyes: Deaf studies talking. University of Minnesota Press. ISBN 978-0-8166-4619-7

Wurtzburg, Susan, and Campbell, Lyle. "North American Indian Sign Language: Evidence for its Existence before European Contact," International Journal of American Linguistics, Vol. 61, No. 2 (Apr., 1995), pp. 153-167.

Kristoffersen, J. H., & Troelsgard, T. (2010, July). The danish sign language dictionary. In Proceedings of the XIV EURALEX International Congress. Leeuwarden: Fryske Akademy (pp. 1549-1554).

Kopf, M., Schulder, M., Hanke, T., & Hénault-Tessier, M. (2021). D6. 1 OVERVIEW OF DATASETS FOR THE SIGN LANGUAGES OF EUROPE

Zhao, L., & Li, S. (2020). Object detection algorithm based on improved YOLOv3. Electronics, 9(3), 537.

Ni, Z., Chen, J., Sang, N., Gao, C., & Liu, L. (2018, October). Light YOLO for high-speed gesture recognition. In 2018 25th IEEE International Conference on Image Processing (ICIP) (pp. 3099-3103). IEEE.

Dima, T. F., & Ahmed, M. E. (2021, July). Using YOLOv5 Algorithm to Detect and Recognize American Sign Language. In 2021 International Conference on Information Technology (ICIT) (pp. 603-607). IEEE.

Mocialov, B., Turner, G., & Hastie, H. (2022). Classification of Phonological Parameters in Sign Languages. arXiv preprint arXiv:2205.12072.

Nimisha, K. P., & Jacob, A. (2020, July). A brief review of the recent trends in sign language recognition. In 2020 International Conference on Communication and Signal Processing (ICCSP) (pp. 186-190). IEEE.

Asri, M. A. M. M., Ahmad, Z., Mohtar, I. A., & Ibrahim, S. (2019). A Real Time Malaysian Sign Language Detection Algorithm Based on YOLOv3. International Journal of Recent Technology and Engineering, 8(2), 651-656.

Kim, S., Ji, Y., & Lee, K. B. (2018, January). An effective sign language learning with object detection based ROI segmentation. In 2018 Second IEEE International Conference on Robotic Computing (IRC) (pp. 330-333). IEEE.

Jocher, G., Stoken, A., Borovec, J., Chaurasia, A., & Changyu, L. (2020). ultralytics/yolov5. Github Repository, YOLOv5.

Wang, C. Y., Liao, H. Y. M., Wu, Y. H., Chen, P. Y., Hsieh, J. W., & Yeh, I. H. (2020). CSPNet: A new backbone that can enhance learning capability of CNN. In Proceedings of the IEEE/CVF conference on computer vision and pattern recognition workshops (pp. 390-391).

## Appendix 1

A plausible modified hybrid architecture is proposed as future work and can be used for the same dataset to experiment for higher accuracy. Such a one stage object detector can be standardised by modifications which is described in details as follows:

*A. Backbone:* The backbone network is responsible for feature extraction and is typically a deep convolutional neural network. In YOLOv5, the backbone network consists of a modified version of EfficientNet, which uses a compound scaling method to balance model size and accuracy. The EfficientNet backbone consists of a series of convolutional blocks, each with depthwise-separable convolutions, followed by a squeeze-and-excitation block that helps to emphasize important features.

*B. Neck:* The neck connects the backbone to the head and is responsible for fusing features at different scales. In YOLOv5, the neck consists of a series of spatial pyramid pooling (SPP) modules, which allow the model to capture features at multiple scales. The SPP modules are followed by a Path Aggregation Network (PAN) that fuses the features from different scales and produces a feature map with a high spatial resolution.

*C. Head:* The head is responsible for predicting bounding boxes and class probabilities. In YOLOv5, the head consists of a series of convolutional layers that process the features from the neck and output bounding boxes and class probabilities. The head also uses anchor boxes to predict the size and aspect ratio of the bounding boxes. YOLOv5 uses a focal loss function that down-weights easy examples and focuses on hard examples during training.

Confidence, Classification and Localisation losses are used in this research paper, however, further four types of losses can be used which are defined as:

A. *Confidence Loss:* This loss function measures the difference between the predicted confidence scores and the ground truth confidence scores. The confidence loss is defined as:

$$L_{conf} = \lambda_{conf} \sum_{i=1}^{S} \sum_{j=1}^{B} [c_i^j - p_i^j]^2$$

where $\lambda_{conf}$ is a hyperparameter that controls the weight of the confidence loss, S is the number of grid cells in the output feature map, B is the number of anchor boxes per grid cell, $c_i^j$ is the ground truth confidence score for anchor box $j$ in grid cell $i$, and $p_i^j$ is the predicted confidence score for anchor box $j$ in grid cell $i$.

B. *Classification Loss:* This loss function measures the difference between the predicted class probabilities and the ground truth class labels. The classification loss is defined as:

$$L_{cls} = \lambda_{cls} \sum_{i=1}^{S} \sum_{j=1}^{B} \sum_{c=1}^{C} [t_i^{j,c} - p_i^{j,c}]^2$$

where $\lambda_{cls}$ is a hyperparameter that controls the weight of the classification loss, $C$ is the number of object classes, $t_i^{j,c}$ is the ground truth class label for anchor box $j$ in grid cell $i$, and $p_i^{j,c}$ is the predicted probability of class $c$ for anchor box $j$ in grid cell $i$.

C. *Localisation Loss:* This loss function measures the difference between the predicted bounding box coordinates and the ground truth bounding box coordinates. The localization loss is defined as:

$$L_{loc} = \lambda_{loc} \sum_{i=1}^{S} \sum_{j=1}^{B} [t_i^j - p_i^j]^2$$

where $\lambda_{loc}$ is a hyperparameter that controls the weight of the localization loss, $t_i^j$ is the ground truth bounding box coordinates for anchor box $j$ in grid cell $i$, and $p_i^j$ is the predicted bounding box coordinates for anchor box $j$ in grid cell $i$.

D. *GIoU Loss:* This loss function is an improved version of the localization loss that takes into account the overlap between predicted and ground truth bounding boxes. The GIoU loss is defined as:

$$L_{giou} = 1 - IoU(b_i, b_i^*) + \lambda_{giou}[g(b_i, b_i^*) - IoU(b_i, b_i^*)]$$

where $b_i$ is the predicted bounding box, $b_i^*$ is the ground truth bounding box, $\lambda_{giou}$ is a hyperparameter that controls the weight of the GIoU loss, IoU is the intersection over union between the two boxes, and $g(b_i, b_i^*)$ is a function that computes the enclosing box of the two boxes.

## Appendix 2

The architecture of the model used in this research paper is presented in the following block diagram (Figure 4).

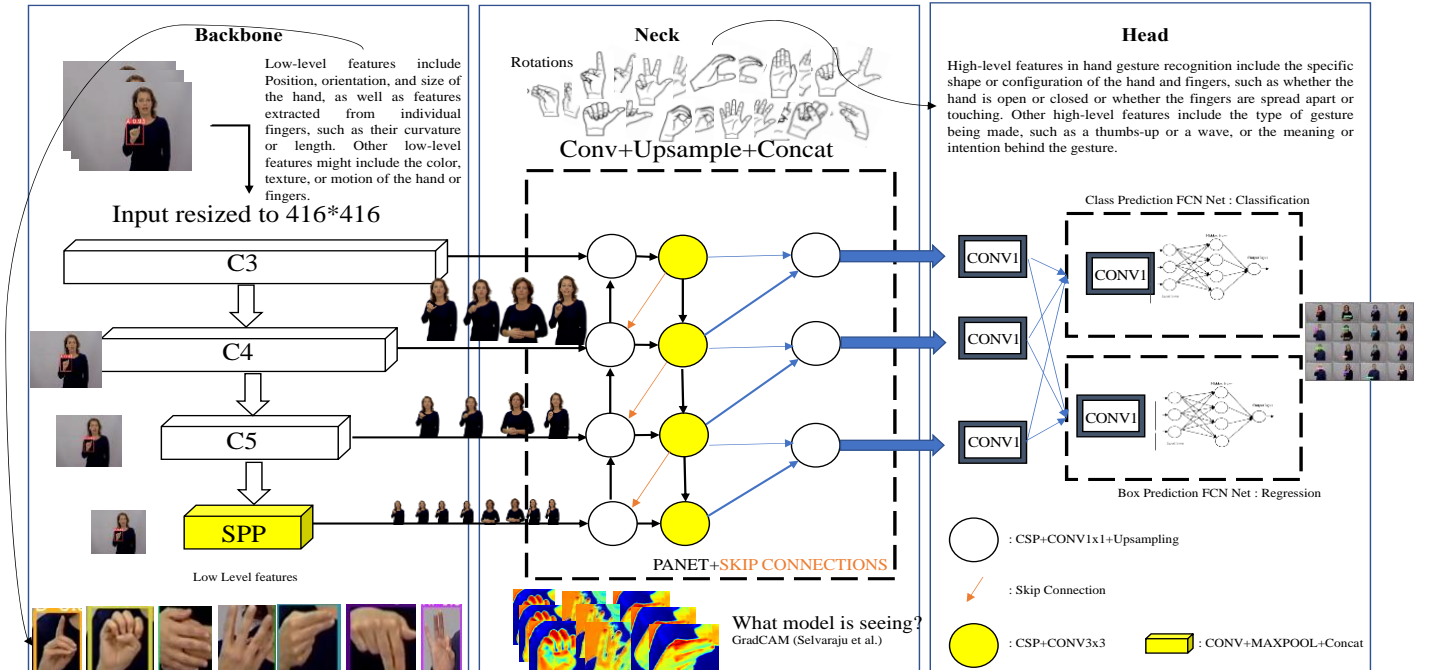

*Figure 4 Block Diagram of Proposed modified YOLO Architecture for Hand Gesture Detection and Sign Language Recognition*